\documentclass[conference]{IEEEtran} 

\IEEEoverridecommandlockouts

\usepackage{cite}
\usepackage{amsmath,amssymb,amsfonts}
\usepackage{algorithmic}
\usepackage{graphicx}
\usepackage{textcomp}
\usepackage{xcolor}
\usepackage{balance}
\usepackage{flushend}
\def\BibTeX{{\rm B\kern-.05em{\sc i\kern-.025em b}\kern-.08em
    T\kern-.1667em\lower.7ex\hbox{E}\kern-.125emX}}

\usepackage{algorithm} 
\usepackage{algorithmic}
\usepackage{url}

\DeclareRobustCommand*{\IEEEauthorrefmark}[1]{%
  \raisebox{0pt}[0pt][0pt]{\textsuperscript{\footnotesize #1}}%
}

\begin{document}

\title{Principal Curvatures Estimation with Applications to Single Cell Data}

\author{\IEEEauthorblockN{Yanlei Zhang\IEEEauthorrefmark{1, 2, $^*$}, Lydia Mezrag\IEEEauthorrefmark{1, 2, $^*$}, Xingzhi Sun\IEEEauthorrefmark{3}, Charles Xu\IEEEauthorrefmark{3, 4}, Kincaid Macdonald\IEEEauthorrefmark{3}, Dhananjay Bhaskar\IEEEauthorrefmark{3,4}, \\ Smita Krishnaswamy\IEEEauthorrefmark{2, 3, 4, 5, $\dag$}, Guy Wolf\IEEEauthorrefmark{1, 2, 7, $\dag$} and Bastian Rieck\IEEEauthorrefmark{6, 7, $\dag$}}

\IEEEauthorblockA{\IEEEauthorrefmark{1}Universit{é} de Montr{é}al, Dept. of Math. \& Stat.; \IEEEauthorrefmark{2}Mila – Quebec AI Institute, Montr{é}al, QC, CA}
\IEEEauthorblockA{\IEEEauthorrefmark{3}Yale University, Dept. of Comp. Sci.; \IEEEauthorrefmark{4}Dept. of Genetics; \IEEEauthorrefmark{5}Applied Mathematics Program, New Haven, CT, USA}

\IEEEauthorblockA{\IEEEauthorrefmark{6}Université de Fribourg, Department of Informatics, Fribourg, FR, CH}
\IEEEauthorblockA{\IEEEauthorrefmark{7}Helmholtz Zentrum München, Institute of AI for Health, Munich, BY, DE}
\IEEEauthorblockA{\IEEEauthorrefmark{$*$}Equal contribution; \IEEEauthorrefmark{$\dag$}Co-senior authors.} 
\thanks{This research was partially funded by Mitacs Globalink Research Award IT40964 [L.M.]; Yale -- Boehringer Ingelheim Biomedical Data Science Fellowship [D.B.]; Hightech Agenda Bavaria, Swiss State Secretariat for Education, Research and Innovation [B.R.]; Humboldt Research Fellowship, CIFAR AI Chair, NSERC Discovery grant 03267, FRQNT grant 343567 [G.W.]; CRM-Simons visiting professor award, NSF career grant 2047856 [S.K.]; and NSF grant DMS-2327211 [G.W., S.K.]. The content provided here is solely the responsibility of the authors and does not necessarily represent the official views of the funding agencies. Correspondence to guy.wolf@umontreal.ca and smita.krishnaswamy@yale.edu}
}


\maketitle

\begin{abstract}
The rapidly growing field of single-cell transcriptomic sequencing (scRNAseq) presents challenges for data analysis due to its massive datasets. A common method in manifold learning consists in hypothesizing that datasets lie on a lower dimensional manifold. This allows to study the geometry of point clouds by extracting meaningful descriptors like curvature. In this work, we will present \textit{Adaptive Local PCA (AdaL-PCA)}, a data-driven method for accurately estimating various notions of intrinsic curvature on data manifolds, in particular principal curvatures for surfaces. The model relies on local PCA to estimate the tangent spaces. The evaluation of AdaL-PCA on sampled surfaces shows state-of-the-art results. Combined with a PHATE embedding, the model applied to single-cell RNA sequencing data allows us to identify key variations in the cellular differentiation.   
\end{abstract}


\section{Introduction}
The estimation of principal curvatures and principal directions is crucial in uncovering directional changes within data manifolds. Indeed, the mean and Gaussian curvatures of surfaces have been studied for several decades in computer graphics and some related areas (e.g., \cite{xin2010,Surazhsky2003, Flynn1989, Zhao1996, rusinkiewicz2004estimating}). Recent methods have proposed to estimate curvature over data manifolds derived from point-cloud data via manifold learning techniques (e.g., \cite{Bella2023, Hou2023, He2020}). However, achieving precision is challenging given the variations in data density and the necessity for high-quality samplings. To address this, various methods have been developed. Volume-based approaches like diffusion curvature by \cite{bhaskar2022diffusion} and \cite{hickok2023intrinsic} heavily depend on accurate distance estimations. Laplace–Beltrami operator-based approaches, as explored in \cite{Sritharan2021} and \cite{bhaskar2022diffusion} encounter limitations in accurately estimating curvature from small sample sizes. Second Fundamental Form-based approaches, as proposed in \cite{Sritharan2021}, demonstrate relatively high-quality curvature estimation for scalar curvature. However, they rely on fixed parameters for neighborhood selection. We introduce adaptability into the estimation process, addressing the challenges associated with variable data density and the absence of intrinsic curvature information by dynamically adjusting parameters based on the local properties of the manifold. This ensures robustness across diverse manifolds. Our \textbf{main contributions} are:

\begin{itemize}

\item We estimate the point-wise Gaussian curvature of point clouds and their underlying principal curvatures, i.e. \textit{How much} the data curves and \textit{in which directions} it curves the most.
\item We dynamically adjust neighborhood scales for local PCA and curvature estimation based on the explained variance ratio. This ensures accurate predictions \textit{without requiring hand-tuning} of the parameters.
\item We demonstrate the fidelity of our method relative to ground truth (Gaussian and mean) curvatures on canonical 2-manifolds. 
\item We illustrate its application to single-cell data analysis, where principal curvatures suggest the directions of cell differentiation.
\end{itemize}

\section{Methods}
For differential geometry preliminaries, we refer the reader to \cite{carmo2016}, as an extensive introduction to this topic would be beyond the scope of this work and its succint presentation. 
\subsection{Local PCA}
Our method starts with Local PCA as described in \cite{singer2012vector}. Given a point cloud $x_1, \cdots, x_m$, we select a neighborhood  $\mathcal{N}_{x_{i},\epsilon_{\text{PCA}}}:= \{x_j : 0 < \| x_j - x_i \| < \epsilon_\text{PCA} \}$ around each point $x_i$ for a hyperparameter $\epsilon_{\text{PCA}} > 0$ that has to be determined. Each data matrix containing the neighbors of a point $x_i$ is shifted to be centered around $x_i$ to get a matrix $X_i = \left[ x_{i_1}-x_i, \ldots, x_{i_{N_i}} - x_i \right] $ where $N_i := |\mathcal{N}_{x_i, \epsilon_\text{PCA}}|$. Then, the columns of $X_i$ are rescaled to $B_i = X_iD_i$ by applying a diagonal weighting matrix $D_i$ to emphasize the importance of local data. Finally, SVD decomposition yields a numerical approximation of the tangent plane. Since, we are interested mainly in surfaces, the first two eigenvectors are selected as a basis for the local tangent space and the third one serves as a normal vector to the surface. 

\subsection{Adaptive Local PCA and parameter selection}
AdaL-PCA uses the explained variance ratio for the first two singular values given by
\begin{equation}
    \rho(r):= \frac{\sum_{i = 1}^2 \sigma_i(r)^2}{\sum_{i = 1}^3 \sigma_i(r)^2}
\end{equation}
to select a suitable parameter $\epsilon_{\text{PCA}}$. This ratio describes the fraction of data variance captured by the tangent plane approximated by the span of the first two singular vectors. We set a threshold $\gamma$ for the ratio $\rho(r)$ and compute the largest $r$-neighborhood that explains a fraction $\gamma$ of the data variance. That is,     
\begin{equation}
    \epsilon_\text{PCA} := \max \left\{ r \ \big| \rho(r) > \gamma \right\}. 
\end{equation} 

\begin{algorithm}[!t]
   \caption{Adaptive Local PCA (AdaL-PCA)}
   \label{alg:adaptpca}
\begin{algorithmic}
   \STATE {\bfseries Input:} Point cloud data $x_1, \ldots, x_m \in \mathbb{R}^3$, query point $p$, kernel function $K$ with supports in [0,1], data bound $\delta$ (maximum pairwise distance in data), ratio bound $\rho_0\in(0, 1)$ for choosing size of PCA neighborhood.
   \FOR{$r \in (0, 0.2\delta]$}
   \STATE $(\mathcal{N}_{p, r}, \mathbf{D_r}) \gets \{ (q, \|q - p \|): 0 < \|q - p \| < r \}$
   \STATE $\mathbf{X} \gets \mathcal{N}_{p, r} - p$ 
   \STATE $\mathbf{D}\gets \text{diag}(\sqrt{K(\mathbf{D_r}/ r)})$
   \STATE $\mathbf{B}\gets \mathbf{D} \mathbf{X}$
   \STATE $\mathbf{U} \mathbf{\Sigma} \mathbf{V}^T \gets \text{SVD}(\mathbf{B})$
   \STATE $ \rho(r) \gets \left\{\sum_{i=1}^{2} \sigma_i^2\big/\sum_{i=1}^{3} \sigma_i^2: \sigma_i \in \mathbf{\Sigma}\right\} $
   \ENDFOR
   \STATE $\epsilon_{\text{PCA}} \gets \max(\{r: \rho(r) > \rho_0\})$
   \STATE $\tau \gets \text{argmin}_r\{\rho(r)\}$
   \STATE {\bfseries Return:} $\epsilon_{\text{PCA}}$, $\tau$
\end{algorithmic}
\end{algorithm}
We use a similar method to select a radius $\tau_{i}$ for estimating the curvature around each data point $x_i$. In this case, we need a neighborhood large enough to capture the ``bending'' of the surface. This is done by computing the lowest value reached by the graph of the explained variance ratio $\rho(r)$,  
\begin{equation}
    \tau := \arg\min_r \left\{\rho(r) \right\}. 
\end{equation}

\begin{figure}[h]
\begin{center}
\centering
\includegraphics[width=\columnwidth]{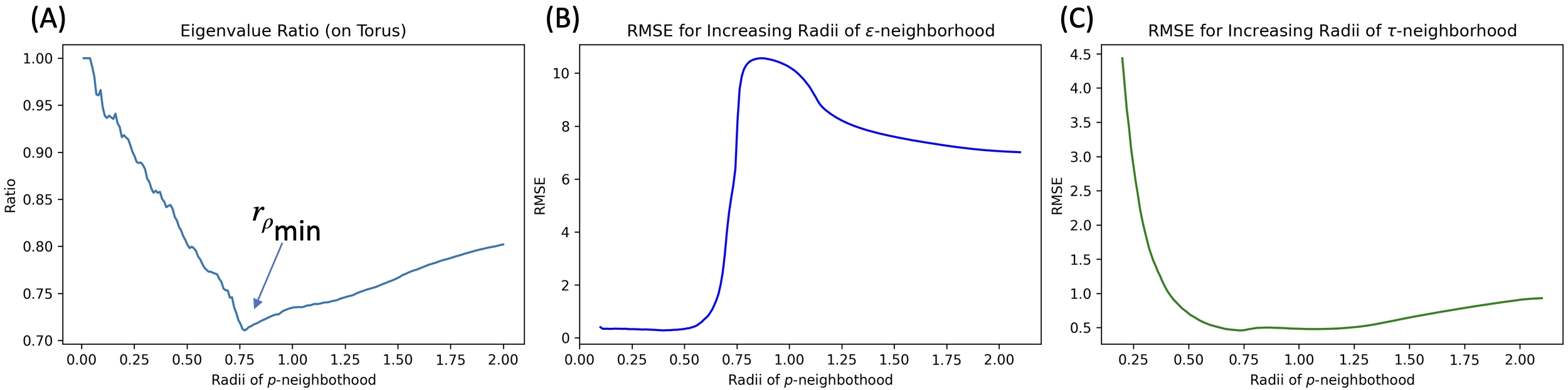}
\caption{\textit{Comparison of the explained variance ratio of the top two singular values and accuracy (RMSE) of Gaussian curvature estimation w.r.t. increasing radii of $\epsilon$-neighborhood and $\tau$-neighborhood around $p$ on torus.}}
\label{fig:epsilon-tau}
\end{center}
\end{figure}

\begin{figure}[!t]
\vskip 0.2in
\begin{center}
\centerline{\includegraphics[width=\columnwidth]{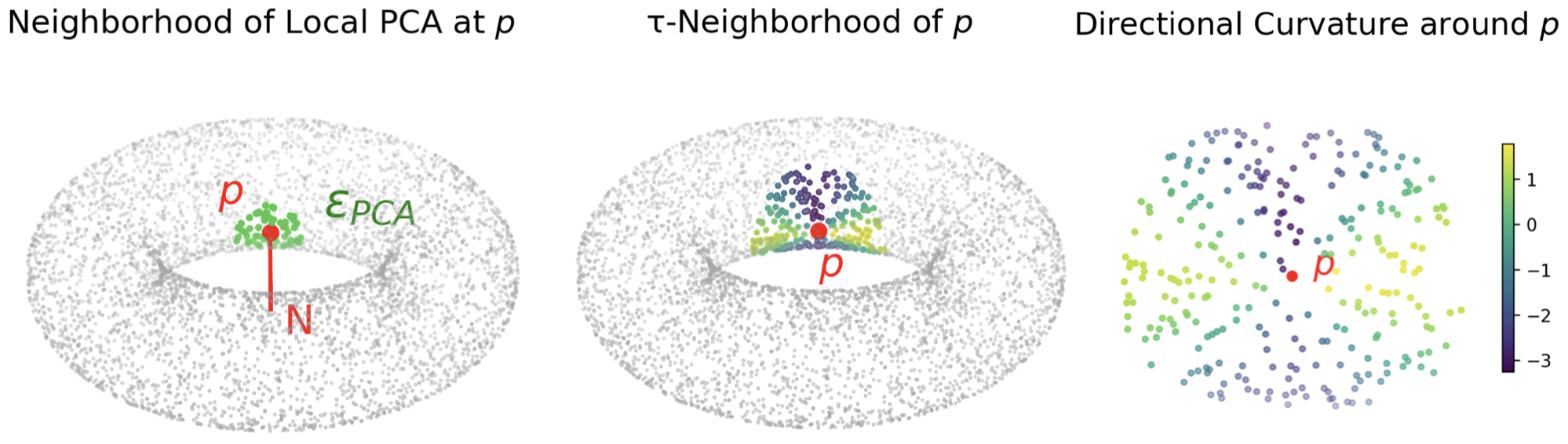}}
\caption{\textit{Directional curvatures in an $\epsilon$-PCA neighborhood of $p$.}}
\label{fig:directional_cur}
\end{center}
\end{figure}
As illustrated in Fig.\ref{fig:epsilon-tau} at a point $p$ on a torus, this approach is motivated by the fact that as the $\tau$-neighborhood increases past a certain threshold, the variance in data can no longer be explained by the selected tangent plane. We refer the reader to Algorithm \ref{alg:adaptpca} for a summary of AdaL-PCA's key steps and emphasize that both $\epsilon_{\text{PCA}}$ and $\tau$ are adjusted at each data point to capture the local geometry. 

\subsection{Curvature Estimation}

\begin{algorithm}[!t]
   \caption{Estimation for Principal Curvature, Gaussian Curvature, and Mean Curvature}
   \label{alg:curvature}
\begin{algorithmic}
   \STATE {\bfseries Input:} Point cloud data $x_1, \ldots, x_m \in \mathbb{R}^3$, query point $p$, kernel function $K$ with supports in [0,1], the pair $(\epsilon_{\text{PCA}}, \tau)$, the percentage $p\in(0, 1)$ of total number of points for which the largest (smallest) directional curvature $\kappa_1$ ($\kappa_2$) is computed.

   \STATE $(\mathcal{N}_{p, \epsilon_{\text{PCA}}}, \mathbf{D_{\epsilon_{\text{PCA}}}}) \gets \{ (q, \|q - p \|): 0 < \|q - p \| < \epsilon_{\text{PCA}}\}$
   \STATE $\mathbf{X} \gets \mathcal{N}_{p, \epsilon_{\text{PCA}}} - p$ 
   \STATE $\mathbf{D}\gets \text{diag}(\sqrt{K(\mathbf{D_{\epsilon_{\text{PCA}}}}/ \epsilon_{\text{PCA}})})$
   \STATE $\mathbf{B}\gets \mathbf{D} \mathbf{X}$
   \STATE $\mathbf{U} \mathbf{\Sigma} \mathbf{V}^T \gets \text{SVD}(\mathbf{B})$
   \STATE $\mathbf{O} \gets \mathbf{U}[\ :3, :\ ]$

   \STATE $\mathcal{N}_{p, \tau}\gets \{ q: 0 < \|q - p \| < \tau \}$
   \FOR{$q\in \mathcal{N}_{p, \tau}$}
   \STATE $v_q \gets q - p$ 
   \STATE $\kappa_q \gets 2(\mathbf{O}[2] \cdot v_q) / ||v_q||^2 $
   \COMMENT{by equation \ref{eq:normal curvature}}
   \STATE $w_q\gets K(v_q/ \tau)$
   \ENDFOR
   \STATE $\text{C} \gets \text{sort} \{(\kappa_q, w_q): \kappa_q \text{ in ascending order}\}$
   \STATE $k \gets \text{int}(p \cdot \text{len}(\text{C}))$
   \STATE $\kappa_1 \gets \text{sum}(\{ \kappa_q \cdot w_q : (\kappa_q, w_q) \in \text{C}[: k]\}) / \text{sum}(\{w_q\})$
   \STATE $\kappa_2 \gets \text{sum}(\{ \kappa_q \cdot w_q : (\kappa_q, w_q) \in \text{C}[-k:]\}) / \text{sum}(\{w_q\})$
   \STATE $K_p \gets \kappa_1 \cdot \kappa_2$
   \STATE $H_p \gets \kappa_1 + \kappa_2$
   \STATE {\bfseries Return:} $\kappa_1$, $\kappa_2$, $K_p$, $H_p$
\end{algorithmic}
\end{algorithm}

The directional curvatures $\kappa_i(T)$ at a point $x_i$ in a direction $T$ are approximated (see for instance \cite{taubin1995estimating}) by 
\begin{equation} \label{eq:normal curvature}
    \kappa_{i}(T) \approx \frac{2N . T}{\lVert T \rVert^{2}} + O(t).    
\end{equation}
Here $T$ is replaced by the entries of $X_i$ in the proper $\tau_i$-neighborhood and $N$ is the orthonormal vector to the frame obtained by local PCA. The principal curvatures $\kappa_1$ and $\kappa_2$ correspond, respectively, to the highest and lowest values of the directional curvatures.\footnote{Note that this is sometimes taken as a definition of principal curvatures.} In practice, we select a percentage (20\%) of the highest (respectively, lowest) curvatures and average them (using a Gaussian kernel) to approximate $\kappa_1$ (respectively, $\kappa_2$). By selecting the directional vectors $T$ corresponding to the highest curvatures $\kappa_T$, this averaging yields principal directions, while we obtain  Gaussian curvature by the product $\kappa_1 \kappa_2$. The time complexity of our current implementation is $O(n_{\tau} m (m^2+ \text{log } n_{\tau}) )$, where $n_{\tau}$ is an upper bound on the cardinality of the $\tau$-neighborhoods (in general $n_{\tau} \ll m$). This can be improved significantly in practice with fast PCA algorithms for scalability \cite{shishkin2019fast},  \cite{xu2023fast}.    

\section{Results and discussion}

Our main contribution is the estimation of the principal curvatures and principal directions. We mainly focus on the application of the principal curvatures and principal directions to biological data and identify key properties and changes in the geometry of these datasets. 

We validate the accuracy of our principal curvature estimation by computing Gaussian curvature on toy datasets. Moreover, we apply our estimation of principal curvature and Gaussian curvature for single-cell RNA sequencing data (scRNA-seq).\footnote{Implementation details and some examples can be found at \url{https://github.com/LydiaMez/AdaL-PCA.git}.} Gaussian curvature gives the ``intensity" for the differentiation of cell states, and principal directions give the directions for the split of the cell lineages. 

\subsection{Estimation on Sampled Surfaces} 

We compare AdaL-PCA's estimates of Gaussian curvature against two contemporary methods, Hickok \& Blumberg \cite{hickok2023intrinsic} and Diffusion Curvature \cite{bhaskar2022diffusion}. We also quantify AdaL-PCA's recovery of ground-truth mean curvature as a validation of the fidelity of its principal curvatures. 

\begin{figure}[!t]
\centering
\includegraphics[width=\columnwidth]{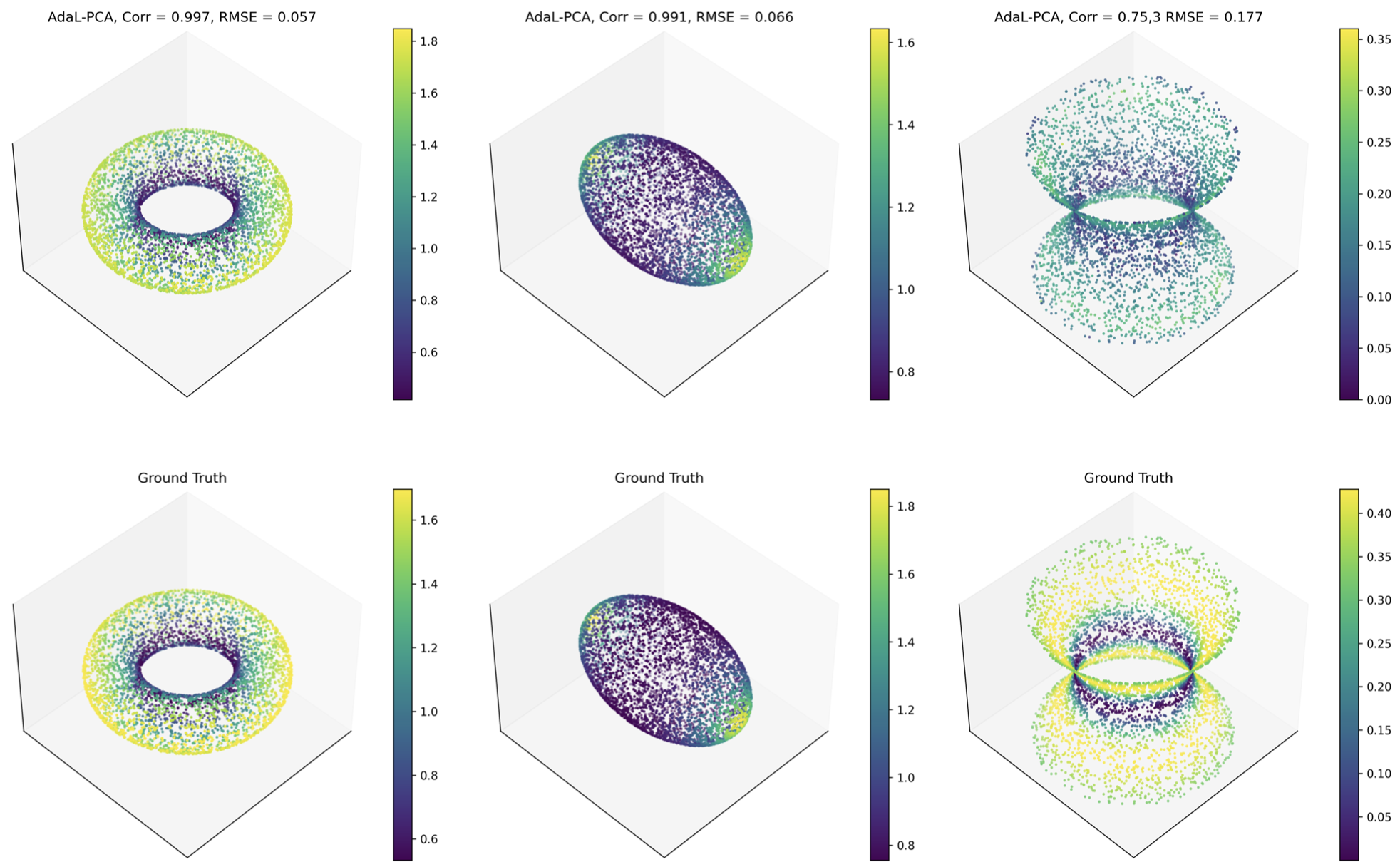}
\caption{\textit{Comparison of AdaL-PCA against ground truth for mean curvature on three toy datasets. Corr stands for Pearson correlation and RMSE stands for the root means squared error.}}
\label{fig:mean_cur}
\end{figure}

To assess the ability of various models to recover the Gaussian and mean curvatures, we generate datasets from three canonical 2-dimensional manifolds: the torus, ellipsoid, and the hyperbolic paraboloid (saddle). Tables \ref{tab:RMSE_EngDist} and \ref{tab:Pearson} were generated from 5000 points sampled uniformly from these surfaces. To study the robustness of each method to noise, we corrupt each point:
\begin{equation*}
    \tilde{x_i} = x_i + \epsilon_i
\end{equation*}
where each $\epsilon_1, \ldots \epsilon_N \overset{\mathrm{iid}}{\sim} \mathcal{N}(0, \sigma)$ and $\sigma $ between 0.0 and 0.5.  

\begin{table}[h]
\renewcommand{\arraystretch}{1.3}
\caption{\textit{Root Mean Square Error (RMSE) and Energy Distance (Eng. Dist) of Gaussian curvature estimation for different noise levels. For RMSE and Eng. Dist, smaller is better.}}
\label{tab:RMSE_EngDist}
    \centering
    \begin{tabular}{c|c||c c||c c }
    \hline
         data & noise & Ours& & H. 
\& B.& \\
          & & RMSE &Eng.Dist& RMSE & Eng.Dist   \\
          \hline  \hline 
         & 0.0 & \bf{0.462}& \bf{0.462} & 1.302 & 0.646  \\ 
         & 0.1 & \bf{1.391}& \bf{0.725} & 7.489 & 3.076  \\
         Torus & 0.2 & \bf{2.023}& \bf{1.026} & 15.914 & 4.772 \\
         & 0.3 & \bf{2.056}& \bf{1.071} & 19.143 & 5.171 \\
         & 0.4 & \bf{2.060}& \bf{1.076} & 19.971 & 5.168 \\
         & 0.5 & \bf{2.048}& \bf{1.059} & 19.944 & 5.041 \\
         \hline 
         & 0.0 & 0.430 & \bf{0.251} & \bf{0.388} & 0.361 \\
         & 0.1 & \bf{0.849}& \bf{0.277} & 6.730 & 3.407  \\
         Ellipsoid & 0.2 & \bf{0.564}& \bf{0.832} & 15.647 & 5.075 \\
         & 0.3 & \bf{1.760}& \bf{1.296} & 20.135 & 5.576 \\
         & 0.4 & \bf{1.988}& \bf{1.565} & 21.007 & 5.541 \\
         & 0.5 & \bf{2.061}& \bf{1.643} & 20.852 & 5.391 \\
         \hline
         & 0.0 & \bf{0.293}& 0.321 & 2.025 & 1.154\\
         & 0.1 & \bf{0.400}& \bf{0.405} & 4.032 & 1.981 \\
         Hyperbolic & 0.2 & \bf{0.567}& \bf{0.538} & 10.077 & 3.588 \\
         paraboloid & 0.3 & \bf{0.673}& \bf{0.674} & 12.829 & 4.065 \\
         & 0.4 & \bf{0.753}& \bf{0.757} & 13.532 & 4.069 \\
         & 0.5 & \bf{0.776}& \bf{0.787} & 13.230 & 3.908 \\
         \hline
    \end{tabular}
\end{table}
Note that both Hickok \& Blumberg's method and Diffusion Curvature require manually specified parameters, which must be tuned \textit{for each} dataset. By contrast, AdaL-PCA's heuristics adapt the method to each dataset.
This results in an improved performance observed in Fig. \ref{tab:RMSE_EngDist} and Fig. \ref{tab:Pearson}. 
Diffusion Curvature is an unsigned measure of local curvature for point clouds sampled from a manifold. Although it differs from Gaussian curvature, numerical experiments detailed in \cite{bhaskar2022diffusion} suggest a correlation. Therefore, we report only the Pearson correlation for diffusion curvature.
\begin{table}[t!]
\centering
\renewcommand{\arraystretch}{1.3}
\caption{\textit{Pearson Correlation Coefficient (Pearson Corr.) of Gaussian curvature estimation for different noise levels.}}
\label{tab:Pearson}
    \begin{tabular}{c|c||c||c}
        \hline
        data & noise & Ours & Diffusion Curvature\\
        \hline \hline
        & 0.0 & \bf{0.996} & 0.445 \\
        & 0.1 & \bf{0.865} & 0.270 \\
      Torus & 0.2 & \bf{0.633} & 0.304 \\
        & 0.3 & \bf{0.550} & 0.308 \\
        & 0.4 & \bf{0.440} & 0.273 \\
        & 0.5 & \bf{0.408} & 0.243 \\
        \hline
        & 0.0 & \bf{0.988} & 0.149 \\
        & 0.1 & \bf{0.325} & 0.057 \\
       Ellipsoid & 0.2 & \bf{0.124} & 0.002 \\
        & 0.3 & -0.153 & \bf{0.017} \\
        & 0.4 & -0.131 & \bf{0.044} \\
        & 0.5 & -0.018 & \bf{0.048} \\
        \hline
        & 0.0 & \bf{0.747} & 0.398 \\
        & 0.1 & \bf{0.603} & 0.333 \\
        Hyperbolic & 0.2 & \bf{0.481} & 0.282 \\
        paraboloid & 0.3 & \bf{0.428} & 0.336 \\
        & 0.4 & \bf{0.386} & 0.342 \\
        & 0.5 & \bf{0.363} & 0.327 \\
        \hline
    \end{tabular}
\end{table}

\subsection{Curvature estimation for single-cell data}

\begin{figure}[t!] 
\centering
\includegraphics[width=\columnwidth]{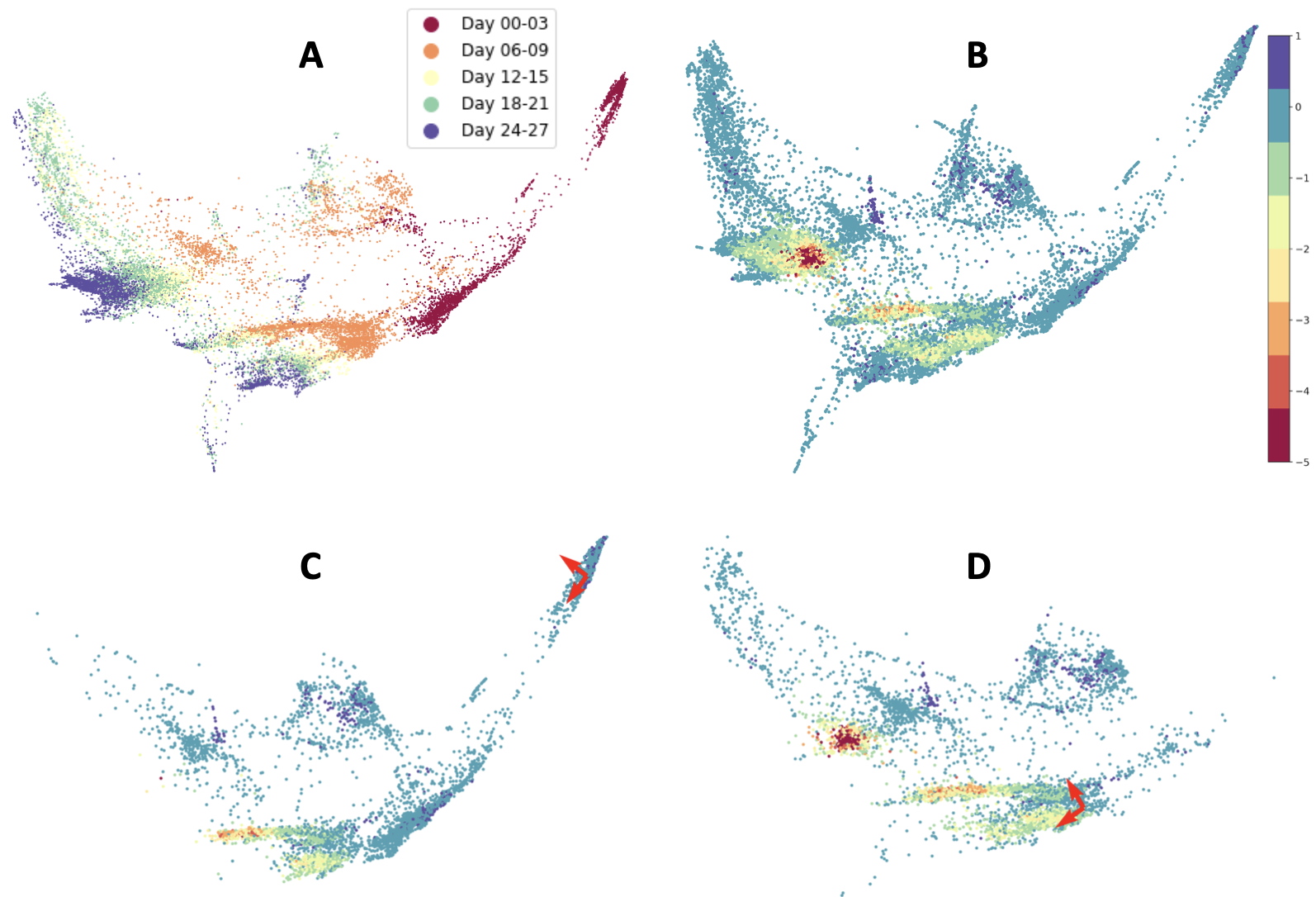}
\caption{\small \textit{Gaussian curvature and principal directions of embryonic stem cell differentiation. \textbf{(A)} PHATE visualization of scRNA-seq data color-coded by time intervals. \textbf{(B)} PHATE plot colored by Gaussian curvature values. \textbf{(C, D)}} Principal directions at different stages of development of cells.}
\label{fig: PHATE}
\end{figure}

\begin{figure}[t!]
\centering
\includegraphics[width=\columnwidth]{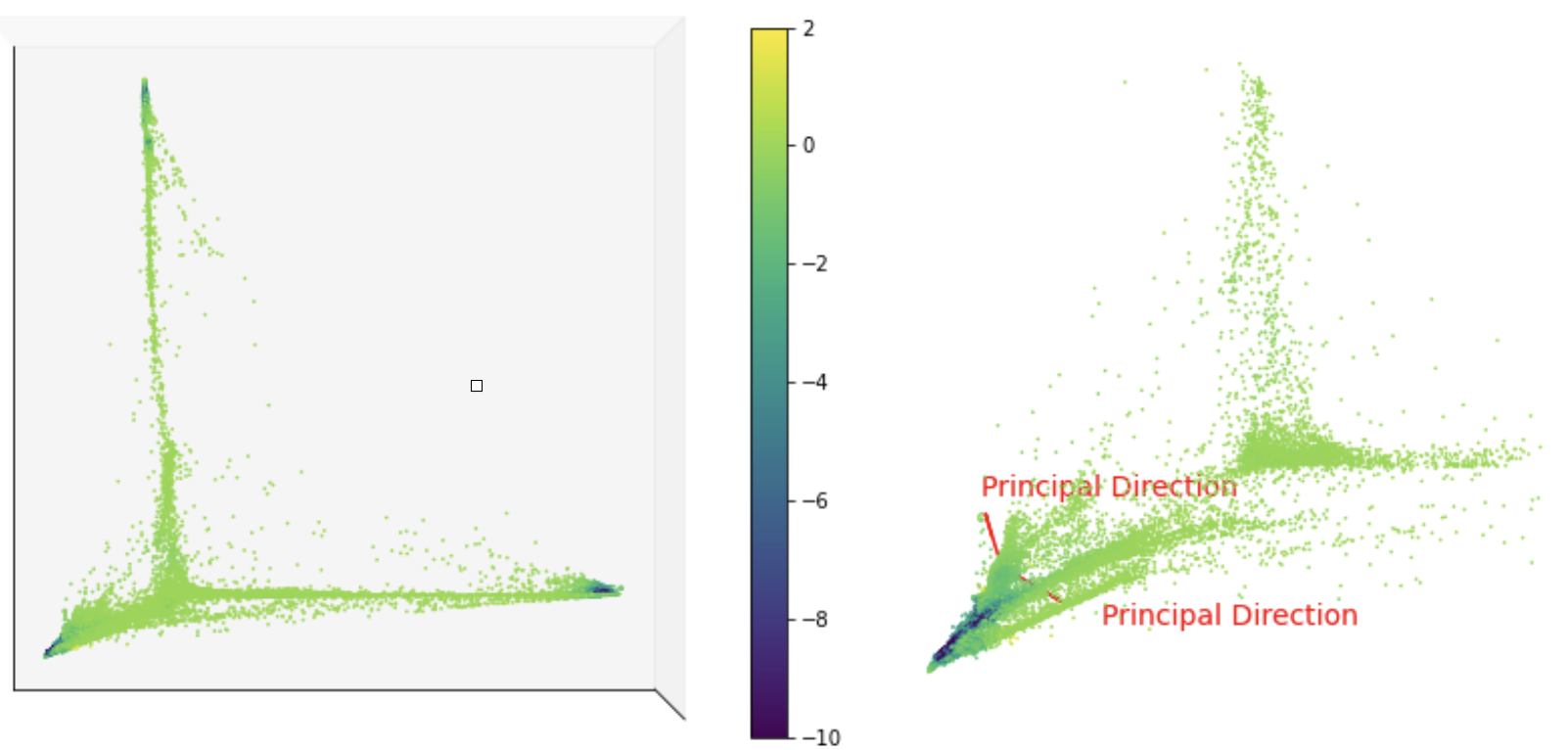}
\caption{\small \textit{Gaussian curvature and principal directions on IPSC dataset using AdaL-PCA.}}
\label{fig:ipsc_cur}
\end{figure}

We apply our model to single-cell data for cell state differentiation direction discovery. We use RNA sequencing data for human embryonic stem cells available at \cite{moon2018embryoid}, collected over 27 days during which cells start as embryonic stem cells and then progressively differentiate into different cellular lineages. Low-dimensional manifold visualization of this data using PHATE (Fig. \ref{fig: PHATE}, A) shows that embryonic cells (days 0-3, displayed in red) branch into two lineages: endoderm (upper split) and ectoderm (lower split) around day 6. Further differentiation occurs during days 12-27. This is reflected in Fig. \ref{fig: PHATE}B with relatively constant zero curvature values at days 0-3 and a transition into a region of high variations in curvature. We observe starting from day 3 a transition into very low negative values of curvature and then a rapid progression into higher values close to zero as we approach day 27. This is consistent with the fact that the region 0-3 days corresponds to the stem state and the region 12-27 to the differentiated state. In addition to the signed curvature that provides a better appreciation of the cellular differentiation into several lineages (cell types), the principal directions in Fig. \ref{fig: PHATE}C, D obtained from projecting the three-dimensional principal directions using the PHATE embedding allow us to track the state towards which the cells differentiate, adding directional information.             

We estimated the curvature of a publicly available single-cell induced pluripotent stem cell (iPSC) reprogramming. In this dataset, mass cytometry is used to quantitatively measure 33 protein biomarkers in 2005 mouse fibroblast cells induced to undergo reprogramming into stem cell state. Low-dimensional PHATE visualization of this data shows fibroblasts progressing to a point of divergence where two lineages emerge, one that successfully undergoes reprogramming and another that undergoes apoptosis (cell death). Our model correctly identifies the initial branching point as having negative values of Gaussian curvature indicating saddle-like divergent paths out of the branching point (Fig. \ref{fig:ipsc_cur}). Moreover, the principal directions on the diverging branch correctly identify the directions in which the cell lineages diverge.

\section{Conclusion} 

We introduced Adaptive Local PCA (AdaL-PCA), a novel method for estimating intrinsic curvature on data manifolds, with a focus on principal curvatures and directions. By dynamically adjusting neighborhood scales based on the explained variance ratio, AdaL-PCA provides robust and accurate curvature estimates without requiring manual parameter tuning. This adaptability effectively handles variations in data density and the lack of prior curvature information, making it ideal for complex, diverse datasets. We validated AdaL-PCA on synthetic surfaces, demonstrating its ability to recover Gaussian and mean curvatures even in noisy settings. Additionally, we applied it to human embryonic single-cell RNA sequencing data, revealing key directions of cellular differentiation and providing biologically meaningful insights. These results highlight AdaL-PCA’s potential in both geometric data analysis and practical applications like single-cell studies. Future work may extend the method to higher dimensions for scalar curvature estimation and improve efficiency by integrating neural network-based local distribution estimation or exploring alternative local PCA frameworks.

\bibliography{ref2}
\bibliographystyle{IEEEtran}

\end{document}